\documentclass{article}

\pdfoutput=1

\DeclareUnicodeCharacter{00A0}{ }  
\DeclareUnicodeCharacter{202F}{ }  
\DeclareUnicodeCharacter{200B}{}   

\DeclareUnicodeCharacter{2212}{-}  
\DeclareUnicodeCharacter{2010}{-}  
\DeclareUnicodeCharacter{2011}{-}  

\DeclareUnicodeCharacter{2013}{--} 
\DeclareUnicodeCharacter{2014}{---}
\DeclareUnicodeCharacter{2019}{'}  
\DeclareUnicodeCharacter{2009}{ }
\DeclareUnicodeCharacter{200A}{ }

\usepackage{arxiv}
\makeatletter
\providecommand{\headeright}{}
\makeatother

\usepackage[utf8]{inputenc} 
\usepackage[T1]{fontenc}    
\usepackage{hyperref}       
\usepackage{url}            
\usepackage{booktabs}       
\usepackage{amsfonts}       
\usepackage{nicefrac}       
\usepackage{amsmath}        
\usepackage{nicefrac}       
\usepackage{microtype}      
\usepackage{textgreek} 
\usepackage{graphicx}
\usepackage{doi}
\usepackage{apacite}
\usepackage{natbib}

\providecommand{\keywords}[1]{\par\noindent\textbf{Keywords: }#1}

\usepackage{titlesec}
\titlespacing{\section}{0pt}{\parskip}{-\parskip}
\titlespacing{\subsection}{0pt}{\parskip}{-\parskip}
\titlespacing{\subsubsection}{0pt}{\parskip}{-\parskip}

\title{A Quantitative Experimental Repeated Measures Study of Training Dynamics in a Small Llama Style Language Model Under a Compute-Aware Token Budget}

\author{ \href{https://orcid.org/0009-0005-5262-1583}{\includegraphics[scale=0.06]{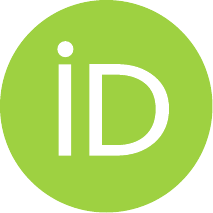}\hspace{1mm}Joe Dwyer}\thanks{Once again, I would like to acknowledge the contributions of Jason M. Pittman. Once again, his advice and knowledge throughout our time working together have been instrumental to my success.} \\
	Department of Computer Information Science\\
	ECPI University\\
	Virginia Beach, VA 23462 \\
	\texttt{jdwyer@ecpi.edu} \\
}
\begin{document}
\maketitle

\begin{abstract}
This study examines training dynamics in a small Llama-style language model trained under a fixed, compute-constrained token budget. Rather than evaluating efficiency solely through endpoint performance, the study uses a quantitative experimental repeated measures design to analyze how validation loss, validation perplexity, rolling volatility, backslide behavior, spike behavior, and between-seed variability change across token-based training intervals. Six independent training runs were conducted on a 4.26-million-parameter model using the TinyStories corpus, CPU-based full-precision training, and a target budget of approximately 20 million cumulative training tokens. Metrics were collected across 21 intervals, producing 126 seed-by-interval observations. Repeated measures ANOVA showed statistically significant interval effects for validation loss, validation perplexity, and rolling volatility. Descriptive trajectories revealed rapid early improvement followed by non-monotonic degradation during later training intervals. Mean validation loss decreased from 8.3552 at initialization to 2.7996 near 4 million tokens, but increased to 3.9010 by the final checkpoint. Validation perplexity followed the same pattern, falling sharply early in training before rising later. Derived telemetry further showed recurrent validation-loss backslides and no interval-summary evidence of a stable phase under the predefined criteria. These findings suggest that compute-aware language model evaluation should examine training trajectories rather than endpoint metrics alone. In constrained compute settings, additional token exposure may increase computational cost without producing proportional generalization gains, and interval-level telemetry can reveal instability, regression, and diminishing returns that final metrics may obscure.

\end{abstract}
\keywords{LLM Training \and Training Dynamics \and repeated measures ANOVA \and Compute Overhead}

\section{Introduction}

Linearly scaled datasets increase compute overhead during large language model training, limiting scalability, reproducibility, and accessibility across the research ecosystem \citep{Chennareddy2024AdaptiveSampling}. Traditional scaling approaches that proportionally increase model size, dataset size, and compute resources exhibit diminishing returns and are increasingly unsustainable as training costs and infrastructure demands continue to rise \citep{Kaplan2020Scaling,Hoffmann2022Chinchilla}. Although compute-optimal scaling strategies have refined these assumptions, much of the existing literature continues to emphasize endpoint performance outcomes rather than the behavior of models during training.

Research in deep learning demonstrates that training dynamics are not linear or monotonic \citep{Belkin2019BiasVariance,Nakkiran2019DeepDoubleDescent}. Validation loss and generalization error can temporarily worsen before improving later in training, reflecting nonlinear learning regimes rather than smooth convergence. These behaviors challenge the assumption that training progress follows a steady trajectory and indicate that instability and regression are inherent features of modern neural network optimization.

Despite these observations, limited empirical attention has been given to how compute-efficient training strategies manifest during training, particularly with respect to stability, variability, and reproducibility across token-based intervals. This gap is especially pronounced in constrained compute environments, where inefficiencies and instability may be amplified and where reproducibility is critical. Without systematic measurement of training behavior across repeated observations, claims of efficiency remain incomplete and difficult to generalize beyond large-scale enterprise training contexts. This study addresses that gap by examining training dynamics under fixed, compute-aware conditions rather than relying solely on endpoint efficiency outcomes.

\section{Purpose of Study}
This study examines training dynamics in a small Llama-style language model trained under a compute-aware, constrained token budget using a fixed corpus. Training stability, non-monotonic learning behavior, and between-seed variability are evaluated across token-based training intervals using a quantitative experimental repeated measures design. Prior work demonstrates that training progress in neural networks is often nonlinear, with periods of regression and instability preceding convergence \citep{Belkin2019BiasVariance,Nakkiran2019DeepDoubleDescent}. By shifting the unit of analysis from endpoint performance outcomes to training behavior observed during optimization, this study reframes compute efficiency as a dynamic process and provides empirical insight into model behavior under constrained compute conditions.

\section{Study Purpose and Research Questions}

This study examines training dynamics in a small Llama-style language model trained under a compute-aware, constrained token budget using a fixed corpus. The study employs a quantitative experimental repeated measures design to evaluate training stability, non-monotonic learning behavior, and between-seed variability across token-based training intervals. The research questions and corresponding hypotheses are summarized in Table~\ref{tab:rq_summary}, along with the associated independent and dependent variables.

\begin{table}[h]
\centering
\caption{Summary of Research Questions, Hypotheses, and Variables}
\label{tab:rq_summary}
\begin{tabular}{p{1.5cm} p{2cm} p{4.5cm} p{5cm}}
\hline
\textbf{RQ} & \textbf{Hypothesis} & \textbf{Independent Variable(s)} & \textbf{Dependent Variable(s)} \\
\hline
RQ1 & H$_{01}$ / H$_{11}$ & 
Cumulative training tokens processed & 
Training stability metrics (e.g., volatility, spike rate, backslide frequency) \\
\hline
RQ2 & H$_{02}$ / H$_{12}$ & 
Random seed & 
Learning trajectory metrics; validation loss and perplexity over training intervals \\
\hline
RQ3 & H$_{03}$ / H$_{13}$ & 
Cumulative training tokens processed & 
Non-monotonic learning behavior metrics (e.g., validation loss spikes, backslides) \\
\hline
RQ4 & H$_{04}$ / H$_{14}$ & 
Cumulative training tokens processed & 
Indicators of a statistically stable training phase derived from stability metrics \\
\hline
\end{tabular}
\end{table}

\subsection{Model, Data, and Training Environment}

The experimental model was a small Llama-style autoregressive language model implemented in PyTorch . The model contained 4,262,144 trainable parameters and used a vocabulary size of 4,096 tokens, a context length of 128 tokens, four transformer layers, four attention heads, an embedding dimension of 256, rotary positional embeddings, and no dropout. This configuration was selected to support repeated experimental runs under a constrained compute environment while preserving the core architectural characteristics of decoder-only transformer language models.

The training corpus was TinyStories, with the training and validation splits held fixed across all experimental runs. A tokenizer with a vocabulary size of 4,096 was trained and reused across runs to ensure consistent tokenization. The experiment used a target token budget of 20,000,000 cumulative training tokens, with evaluation checkpoints scheduled at approximately 1,000,000-token intervals. Because training progressed in fixed token increments determined by batch size, context length, and gradient accumulation, the final logged token count was 20,000,768 tokens.

All experiments were conducted on a CPU-based Windows environment using PyTorch CPU execution. Training was performed using full precision \texttt{fp32}. The training configuration used a micro-batch size of 2, gradient accumulation over 8 micro-batches, and a context length of 128 tokens, resulting in 2,048 training tokens per optimizer step. The learning rate was initialized at 0.0004, decayed toward a minimum learning rate of 0.00004, and used a warmup period of 1,000,000 tokens. Weight decay was set to 0.1, AdamW optimizer coefficients were set to $\beta_1 = 0.9$ and $\beta_2 = 0.95$, and gradient clipping was applied at 1.0.

This experimental configuration represents a compute-constrained follow-up study rather than a large-scale compute-optimal training run. The purpose of this design was not to maximize benchmark performance, but to examine whether training dynamics, instability, and between-seed variability could be measured systematically under a fixed and reproducible token budget.

\subsection{Data Collection and Repeated Measurements}

Training telemetry metrics were recorded at fixed token-based intervals throughout the training process. Metrics were collected at initialization and then at approximately every 1,000,000 training tokens until the 20,000,000-token target was reached. Each interval recorded training loss, validation loss, validation perplexity, elapsed time, tokens processed per second, learning rate, device type, and numerical precision.

Six independent runs were conducted using random seeds 101, 202, 303, 404, 505, and 606. Random seed served as the repeated-measures subject identifier. Each seed was trained under the same model architecture, dataset, optimizer configuration, token budget, and evaluation schedule. This design produced 21 repeated observations per seed, resulting in 126 total seed-by-interval observations.

\subsection{Variables and Measures}

The primary within-subjects independent variable was cumulative training interval, operationalized as the checkpoint index corresponding to cumulative tokens processed. Random seed served as the unit of replication and subject identifier in the repeated measures design.

The primary dependent variables were validation loss and validation perplexity. Additional derived stability measures were computed from validation loss trajectories. These included validation loss change between adjacent checkpoints, rolling volatility, backslide indicators, backslide magnitude, spike indicators, spike rate, backslide frequency, and between-seed validation loss variance.

A backslide was defined as an interval in which validation loss increased relative to the prior checkpoint. Spike detection used a rolling threshold based on recent validation loss deltas, with a minimum spike threshold of 0.01 and a rolling z-threshold of 2.0. Rolling volatility was computed using a five-interval rolling window. A stable phase candidate indicator was computed from rolling volatility, spike rate, and backslide frequency; however, this indicator was treated descriptively rather than as a confirmatory endpoint.

\subsection{Statistical Analysis}

The primary inferential analyses used one-way repeated measures analysis of variance, with training interval treated as the within-subjects factor and random seed treated as the subject. Separate repeated measures ANOVAs were conducted for validation loss, validation perplexity, and rolling volatility. Normality was evaluated using Shapiro--Wilk tests computed at each interval. Sphericity was evaluated using Mauchly's test, and Greenhouse--Geisser corrected values were reported where appropriate.

Pairwise comparisons of validation loss across intervals were conducted using paired tests with Bonferroni correction. Because the purpose of the study was to examine training dynamics rather than endpoint superiority alone, statistical significance was interpreted as evidence that model behavior varied across training intervals. Directional interpretation was based on descriptive trajectories, validation loss trends, perplexity trends, and derived stability measures.

A linear mixed effects model was also fit as a robustness check, with validation loss modeled as a function of training interval and random seed included as the grouping factor. This model was used to confirm that interval-level effects were not artifacts of the repeated measures ANOVA framework alone.

\subsection{Instruments}

Data were collected using software-based experimental instruments. These included the PyTorch training framework, deterministic run configurations, automated logging routines, checkpoint-based validation procedures, and post-training telemetry scripts \citep{dwyer2026_trainingdynamicscode}. The telemetry pipeline aggregated per-seed metric files, derived stability measures from validation loss trajectories, and generated interval-level summaries for statistical analysis.

\subsection{Procedure}

The experimental procedure followed a fixed sequence. First, the dataset and tokenizer were prepared using the predefined TinyStories preprocessing configuration. Next, the model architecture, optimizer settings, token budget, checkpoint schedule, and telemetry settings were loaded from the experiment configuration. For each seed, the model was initialized independently while all other experimental conditions were held constant.

Training proceeded until the target token budget was reached. At each checkpoint interval, the model was evaluated on the validation split and telemetry values were written to structured CSV files. After all seeds completed training, per-seed metrics were aggregated into a combined metrics file. Derived telemetry measures were then computed, including validation loss deltas, rolling volatility, backslide frequency, spike rate, and between-seed variance. Finally, repeated measures ANOVA, pairwise comparisons, and mixed effects robustness checks were conducted using the aggregated telemetry data.

\section{Results}

\subsection{Run Completion and Telemetry Coverage}

The completed experiment produced telemetry for six random seeds across 21 training intervals, yielding 126 total observations. No duplicate seed-by-interval rows were removed during preprocessing. The analysis used the derived telemetry file as the primary input for statistical testing. Each seed completed the full token schedule through the final logged checkpoint of 20,000,768 cumulative training tokens.

\subsection{Descriptive Training Trajectory}

Validation loss changed substantially over the course of training. At initialization, the mean validation loss across seeds was 8.3552, with a standard deviation of 0.0167. After approximately 1,001,472 tokens, mean validation loss decreased sharply to 3.6949. By approximately 4,001,792 tokens, mean validation loss reached 2.7996. The lowest mean validation loss occurred near the early-middle portion of training, with a mean validation loss of 2.7996 at interval 4 and 2.8077 at interval 5.

\begin{figure}[h]
\centering
\includegraphics[width=0.95\linewidth]{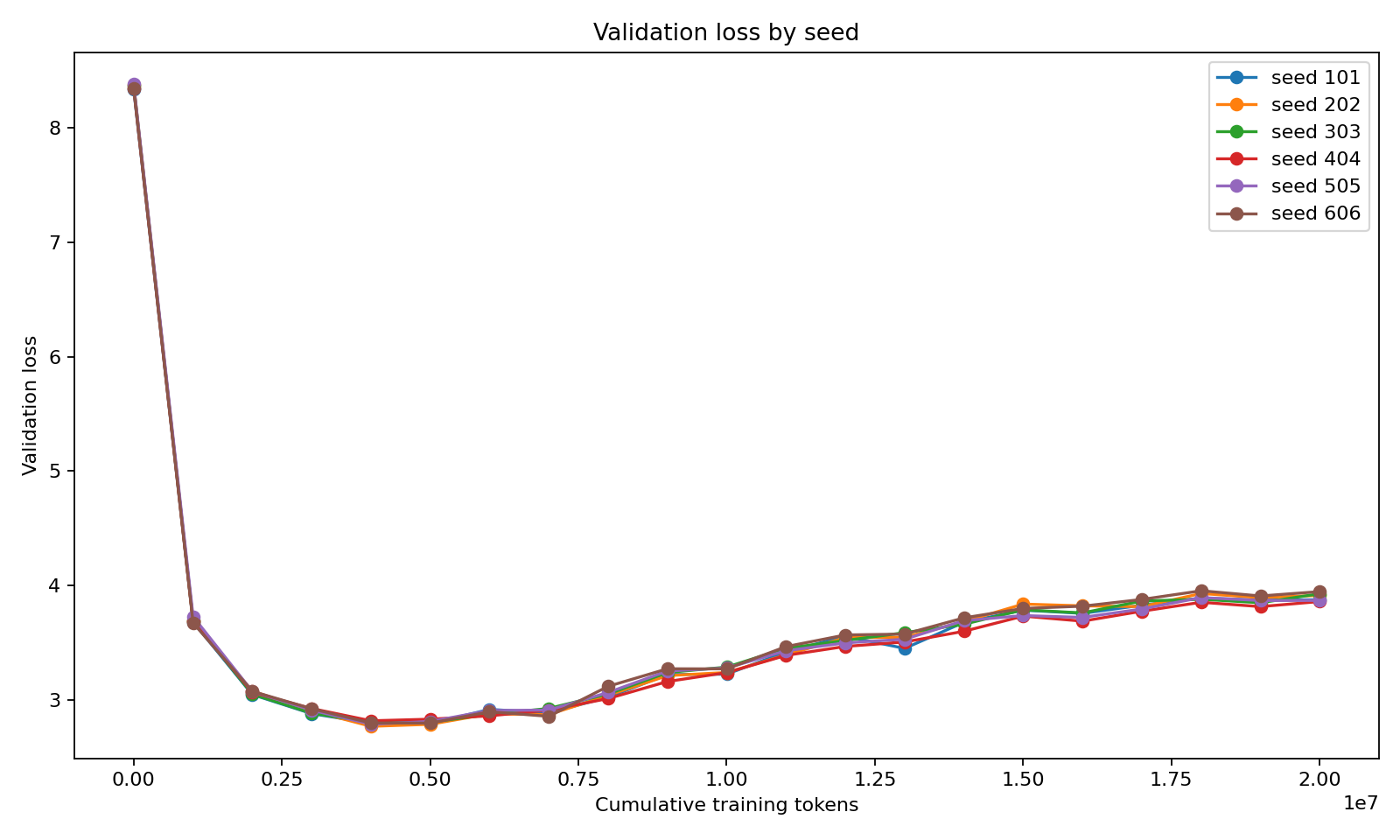}
\caption{Validation loss trajectories by random seed across cumulative training tokens. The trajectories show rapid early improvement followed by non-monotonic degradation during later training intervals.}
\label{fig:validation_loss_by_seed}
\end{figure}

After this early improvement, validation loss became non-monotonic. Mean validation loss increased from 2.8077 at approximately 5,001,216 tokens to 2.8924 at approximately 6,000,640 tokens and continued to rise across later intervals. At the final checkpoint of 20,000,768 tokens, the mean validation loss was 3.9010, with a standard deviation of 0.0355. Mean validation perplexity followed the same general pattern, decreasing from 4252.7436 at initialization to 16.4403 near interval 4, then increasing to 49.4787 at the final checkpoint.

\begin{table}[h]
\centering
\caption{Selected Interval-Level Training Outcomes}
\label{tab:selected_interval_outcomes}
\begin{tabular}{rrrrr}
\toprule
\textbf{Interval} & \textbf{Mean Tokens Seen} & \textbf{Mean Val. Loss} & \textbf{Val. Loss SD} & \textbf{Mean Val. PPL} \cr
0  & 0        & 8.3552 & 0.0167 & 4252.7436 \cr
1  & 1001472  & 3.6949 & 0.0188 & 40.2488 \cr
4  & 4001792  & 2.7996 & 0.0166 & 16.4403 \cr
5  & 5001216  & 2.8077 & 0.0161 & 16.5737 \cr
10 & 10000384 & 3.2597 & 0.0253 & 26.0475 \cr
15 & 15001600 & 3.7822 & 0.0396 & 43.9406 \cr
20 & 20000768 & 3.9010 & 0.0355 & 49.4787 \cr
\bottomrule
\end{tabular}
\end{table}

These results indicate that the model learned rapidly during the earliest training intervals, but additional training tokens did not produce monotonic improvement. Instead, the trajectory showed an initial improvement phase followed by degradation in validation loss and validation perplexity, consistent with non-monotonic training dynamics.

\subsection{Training Stability and Non-Monotonic Behavior}

Derived telemetry metrics provided additional evidence of non-monotonic training behavior. Backslide frequency increased after the early improvement phase. At interval 5, the mean backslide frequency was 0.5. By interval 8, the mean backslide frequency increased to 3.0, and it remained elevated across much of the later training trajectory. Between intervals 11 and 18, mean backslide frequency ranged from 3.8333 to 4.8333, indicating that validation loss increases were recurrent rather than isolated.

\begin{figure}[h]
\centering
\includegraphics[width=0.95\linewidth]{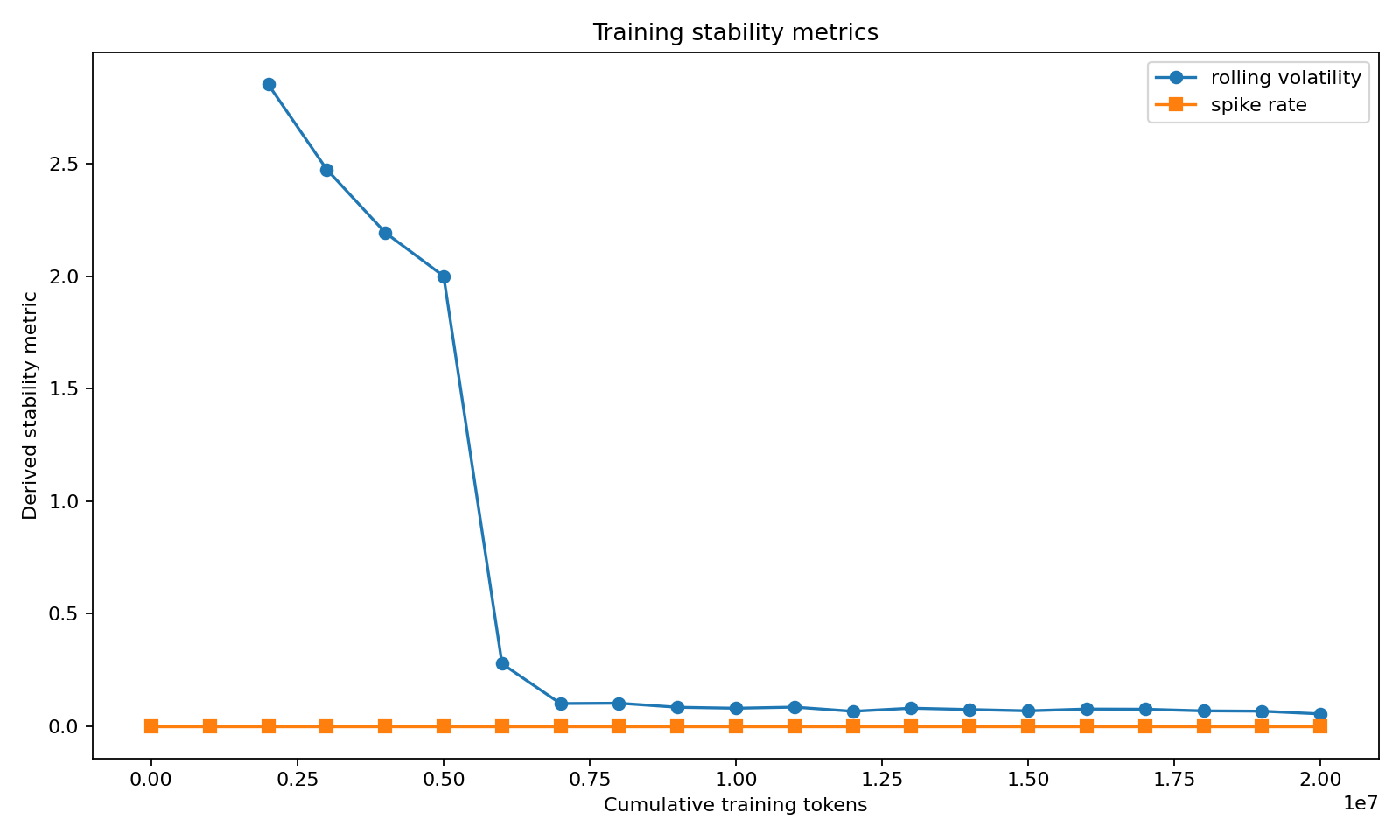}
\caption{Derived training stability metrics across cumulative training tokens. Rolling volatility decreases after the early training phase, while the broader telemetry results indicate continued backslides and a lack of a stable phase under the predefined criteria.}
\label{fig:stability_metrics}
\end{figure}

Rolling volatility was highest during the earliest transition intervals, reflecting the large validation loss changes observed during initial learning. Rolling volatility declined substantially after the early training phase, but this decline did not indicate stable improvement. Instead, later intervals showed lower volatility combined with repeated validation loss backslides and increasing mean validation loss. The stable phase candidate indicator remained 0.0 across all summarized intervals, suggesting that the predefined stability criteria were not met during the observed run.

\begin{table}[h]
\centering
\caption{Selected Derived Stability Metrics by Interval}
\label{tab:stability_metrics}
\begin{tabular}{rrrrr}
\toprule
\textbf{Interval} & \textbf{Mean Tokens Seen} & \textbf{Rolling Volatility} & \textbf{Backslide Frequency} & \textbf{Stable Phase Share} \cr
2  & 2000896  & 2.8520 & 0.0000 & 0.0000 \cr
5  & 5001216  & 1.9997 & 0.5000 & 0.0000 \cr
8  & 8001536  & 0.1030 & 3.0000 & 0.0000 \cr
12 & 12001280 & 0.0671 & 4.8333 & 0.0000 \cr
15 & 15001600 & 0.0691 & 4.8333 & 0.0000 \cr
18 & 18001920 & 0.0690 & 4.0000 & 0.0000 \cr
20 & 20000768 & 0.0554 & 3.0000 & 0.0000 \cr
\bottomrule
\end{tabular}
\end{table}

The stability results suggest that the training process did not settle into a statistically stable improvement phase under the predefined criteria. Rather, the model exhibited an early learning phase followed by persistent non-monotonic behavior.

\subsection{Repeated Measures ANOVA for Validation Loss}

A one-way repeated measures ANOVA was conducted to test whether validation loss differed significantly across training intervals. The effect of interval was statistically significant, $F(20, 100) = 10030.87$, $p < .001$. The Greenhouse--Geisser corrected p-value also remained statistically significant, $p < .001$. The generalized eta-squared effect size was $\eta^2_g = .9993$, indicating that the training interval accounted for nearly all explainable variation in validation loss under this repeated measures design.

This result supports rejection of the null hypothesis that validation loss remained constant across cumulative training intervals. However, the direction of change was not uniformly beneficial. Descriptive trajectories showed that validation loss improved sharply during early training and then worsened during later training intervals.

\subsection{Repeated Measures ANOVA for Validation Perplexity}

A repeated measures ANOVA was also conducted for validation perplexity. The effect of interval was statistically significant, $F(20, 100) = 21061.95$, $p < .001$. The Greenhouse--Geisser corrected p-value remained statistically significant, $p < .001$, with $\eta^2_g = .9997$. As with validation loss, this finding indicates that validation perplexity varied significantly across training intervals.

The descriptive perplexity trajectory paralleled validation loss. Mean validation perplexity decreased dramatically during the earliest intervals, reaching its lowest observed mean near interval 4, before increasing during later training. This pattern reinforces the interpretation that continued token exposure under this configuration did not produce monotonic generalization improvement.

\subsection{Repeated Measures ANOVA for Rolling Volatility}

A repeated measures ANOVA was conducted for rolling volatility to examine whether training stability changed across intervals. The effect of interval was statistically significant, $F(18, 90) = 15390.48$, $p < .001$. The Greenhouse--Geisser corrected p-value also remained statistically significant, $p < .001$, with $\eta^2_g = .9995$.

This result indicates that stability behavior changed significantly across the training trajectory. Rolling volatility was largest during the early learning phase, when validation loss changed rapidly, and lower during later intervals. However, lower volatility did not correspond to stable improvement because validation loss backslides persisted, and the final validation loss was higher than the early minimum.

\subsection{Normality, Sphericity, and Robustness Checks}

Shapiro--Wilk tests were computed by interval for validation loss. The minimum observed p-value was 0.1477, suggesting that the interval-level validation loss distributions did not show evidence of severe normality violations under the available six-seed design.

The sphericity diagnostic reported sphericity as satisfied for the validation loss analysis. Nevertheless, Greenhouse--Geisser corrected p-values were reported as a conservative check because repeated measures designs with many intervals and small numbers of subjects can be sensitive to assumption violations.

A linear mixed effects model was fit as a robustness check, using validation loss as the dependent variable, interval as a fixed effect, and seed as the grouping factor. The model converged successfully with 126 observations and 6 groups. The fixed interval effects were statistically significant across the modeled interval contrasts, supporting the repeated measures ANOVA finding that validation loss differed systematically across training intervals.

\subsection{Research Question Summary}

For RQ1, cumulative training tokens were associated with significant changes in training stability metrics. Rolling volatility differed significantly across intervals, and backslide frequency increased after the early learning phase. These findings support rejection of the null hypothesis for RQ1.

For RQ2, the repeated seed design showed broadly similar learning trajectories across seeds, but measurable between-seed variance was present at each interval. Between-seed validation loss standard deviation increased from 0.0167 at initialization to 0.0355 at the final checkpoint, indicating that random seed contributed to modest but observable trajectory variation.

For RQ3, the results provide clear evidence of non-monotonic learning behavior. Validation loss and validation perplexity improved sharply early in training, reached their best mean values around the 4--5 million token range, and then worsened across later intervals. Backslide metrics further confirmed that validation loss increases occurred repeatedly during training.

For RQ4, the predefined stable phase candidate criteria were not met at the interval-summary level. Stable phase share remained 0.0 across the summarized intervals. This finding suggests that, under the observed compute-constrained conditions, the model did not enter a statistically stable training phase as operationalized in this study.

\begin{table}[h]
\centering
\caption{Summary of Findings by Research Question}
\label{tab:results_summary}
\begin{tabular}{p{1.3cm} p{4.2cm} p{5.5cm} p{3cm}}
\hline
\textbf{RQ} & \textbf{Focus} & \textbf{Primary Finding} & \textbf{Conclusion} \\
\hline
RQ1 & 
Training stability across cumulative token intervals & 
Rolling volatility differed significantly across intervals, $F(18, 90) = 15390.48$, $p < .001$, and backslide frequency increased after the early learning phase. & 
Reject H$_{01}$ \\
\hline
RQ2 & 
Between-seed variability in learning trajectories & 
All seeds followed a similar broad trajectory of early improvement followed by later degradation, but measurable between-seed variance was present across intervals. & 
Descriptive evidence of seed-level variability observed \\
\hline
RQ3 & 
Non-monotonic learning behavior across cumulative token intervals & 
Validation loss and perplexity improved sharply early in training, reached their best mean values around the 4--5 million token range, and worsened during later intervals. & 
Reject H$_{03}$ \\
\hline
RQ4 & 
Presence of a statistically stable training phase & 
The predefined stable phase candidate indicator remained 0.0 across summarized intervals, indicating no stable phase under the study criteria. & 
Fail to reject H$_{04}$ \\
\hline
\end{tabular}
\end{table}

\subsection{Summary of Findings}

Overall, the results show that training interval had a statistically significant effect on validation loss, validation perplexity, and rolling volatility. The training trajectory was strongly non-monotonic: rapid early improvement was followed by later degradation in validation loss and perplexity. These findings support the central claim that compute-aware training evaluation should not rely solely on endpoint outcomes. Instead, training dynamics across token intervals reveal important patterns of instability, regression, and variability that would be obscured by a single final metric.

\section{Conclusion}

This study examined training dynamics in a small Llama-style language model trained under a fixed, compute-constrained token budget. Rather than evaluating training solely through endpoint performance, the study measured validation loss, validation perplexity, rolling volatility, backslide behavior, spike behavior, and between-seed variation across repeated token-based intervals. The results demonstrate that training behavior was strongly interval-dependent and non-monotonic. Validation loss and validation perplexity improved rapidly during the earliest training intervals, reached their best observed values during the early-middle portion of training, and then degraded across later intervals.

The findings support the central argument that compute-aware model evaluation should account for training dynamics, not merely final performance outcomes. Under the observed configuration, additional token exposure did not produce steady improvement. Instead, the model displayed recurrent backslides and failed to satisfy the predefined criteria for a stable training phase. This pattern suggests that, in constrained compute settings, more training tokens may increase computational cost without yielding proportional gains in generalization. In some cases, continued training may even move the model away from its best observed validation performance.

The repeated measures design provided a useful framework for identifying these effects. Treating random seed as the repeated-measures subject allowed the study to evaluate within-run changes over time while preserving between-seed replication. The statistically significant effects observed for validation loss, validation perplexity, and rolling volatility indicate that training interval was not a neutral factor. Instead, cumulative token exposure meaningfully shaped model behavior across the trajectory. The mixed effects robustness check further supported the conclusion that interval-level differences were systematic rather than artifacts of a single seed or isolated checkpoint.

These results have practical implications for small-scale and resource-constrained language model experimentation. Researchers and educators working outside large industrial compute environments often cannot rely on massive training runs or repeated large-scale sweeps. For such settings, telemetry-rich training procedures can provide insight into when useful learning occurs, when instability emerges, and when additional computation may no longer be justified. The results of this study suggest that lightweight repeated-measures experiments can reveal meaningful training dynamics even when the model and token budget are modest.

The study also has limitations. The experiment used a small model, a fixed TinyStories corpus, CPU-based training, and a limited number of random seeds. Therefore, the findings should not be interpreted as general claims about all Llama-style models, all datasets, or all compute budgets. The stability metrics used in this study were derived from validation loss trajectories and should be interpreted as operational measures rather than universal definitions of stable learning. In addition, the observed non-monotonic behavior may reflect the interaction among model size, corpus repetition, learning-rate schedule, and the constrained token budget.

Future research should extend this design across larger models, additional datasets, alternative learning-rate schedules, and different token budgets. Additional work should also compare endpoint-based efficiency measures with interval-based training dynamics to determine when early stopping, adaptive sampling, or curriculum-based approaches may reduce unnecessary computation. Expanding the number of seeds and hardware configurations would further clarify how reproducible these dynamics are across experimental conditions.

Overall, this study shows that training efficiency is not merely a question of how many tokens are processed or what final validation metric is achieved. Efficiency also depends on how the model behaves throughout training. By examining the trajectory rather than only the endpoint, this study provides evidence that training dynamics can reveal instability, regression, and diminishing returns that would otherwise remain hidden. For compute-aware language model research, these findings reinforce the importance of measuring not only whether a model improves, but when, how, and for how long that improvement remains stable.

\bibliographystyle{apacite}
\bibliography{references}

@article{Belkin2019BiasVariance,
  author  = {Belkin, Mikhail and Hsu, Daniel and Ma, Siyuan and Mandal, Soumik},
  title   = {Reconciling modern machine learning practice and the bias--variance trade-off},
  journal = {Proceedings of the National Academy of Sciences},
  year    = {2019},
  volume  = {116},
  number  = {32},
  pages   = {15849--15854},
  doi     = {10.1073/pnas.1903070116},
  url     = {https://doi.org/10.1073/pnas.1903070116}
}

@inproceedings{Chennareddy2024AdaptiveSampling,
  author       = {Chennareddy, Vamshi and Patibandla, Praneeth Kumar and Koppula, Raghavendra Chary},
  title        = {A methodology for efficient data processing in {AI} applications using adaptive sampling and automated testing},
  booktitle    = {Proceedings of the 4th International Conference on Computer Communication and Artificial Intelligence (CCAI)},
  year         = {2024},
  pages        = {552--556},
  doi          = {10.1109/CCAI61966.2024.10603318}
}

@article{Hoffmann2022Chinchilla,
  author       = {Hoffmann, Jordan and Borgeaud, Sebastian and Mensch, Arthur and Buchatskaya, Elena and Cai, Trevor and Rutherford, Eliza and de Las Casas, Diego and Hendricks, Lisa Anne and Welbl, Johannes and Clark, Aidan and Hennigan, Tom and Noland, Eric and Millican, Katie and van den Driessche, George and Damoc, Bogdan and Guy, Aurelia and Osindero, Simon and Simonyan, Karen and Elsen, Erich and Rae, Jack W. and Vinyals, Oriol and Sifre, Laurent},
  title        = {Training compute-optimal large language models},
  year         = {2022},
  journal      = {arXiv},
  eprint       = {2203.15556},
  archivePrefix= {arXiv},
  url          = {https://arxiv.org/abs/2203.15556}
}

@article{Kaplan2020Scaling,
  author       = {Kaplan, Jared and McCandlish, Sam and Henighan, Tom and Brown, Tom B. and Chess, Benjamin and Child, Rewon and Gray, Scott and Radford, Alec and Wu, Jeffrey and Amodei, Dario},
  title        = {Scaling laws for neural language models},
  year         = {2020},
  journal      = {arXiv},
  eprint       = {2001.08361},
  archivePrefix= {arXiv},
  url          = {https://arxiv.org/abs/2001.08361}
}

@article{Nakkiran2019DeepDoubleDescent,
  author  = {Nakkiran, Preetum and Kaplun, Gal and Bansal, Yamini and Yang, Trenton and Barak, Boaz and Sutskever, Ilya},
  title   = {Deep double descent},
  journal = {arXiv},
  year    = {2019},
  eprint  = {1912.02292},
  archivePrefix = {arXiv},
  primaryClass  = {cs.LG},
  url     = {https://arxiv.org/abs/1912.02292}
}

@software{dwyer2026_trainingdynamicscode,
  author    = {Dwyer, Joe},
  title     = {TrainingDynamicsCode},
  year      = {2026},
  publisher = {Zenodo},
  doi       = {10.5281/zenodo.20533382},
  url       = {https://doi.org/10.5281/zenodo.20533382}
}
\end{document}